\documentclass[10pt,twocolumn,letterpaper]{article}
\usepackage{amsmath}
\DeclareMathOperator*{\argmin}{argmin}
\usepackage{cvpr}              % To produce the CAMERA-READY version
\usepackage{multirow}
\definecolor{cvprblue}{rgb}{0.21,0.49,0.74}
\usepackage[pagebackref,breaklinks,colorlinks,allcolors=cvprblue]{hyperref}

\def\paperID{4483} % *** Enter the Paper ID here
\def\confName{CVPR}
\def\confYear{2025}

\title{Stable-SCore: A Stable Registration-based Framework \\ for 3D Shape Correspondence}

\author{Haolin Liu\textsuperscript{\rm 1,2,3}$^{*}$, Xiaohang Zhan\textsuperscript{\rm 2}$^{*}$, Zizheng Yan\textsuperscript{\rm 1,4}$^{*}$, Zhongjin Luo\textsuperscript{\rm 1,4}, Yuxin Wen\textsuperscript{\rm 2}, Xiaoguang Han\textsuperscript{\rm 4,1}$^{\dag}$ \\
\small{$^{*}$equal contribution} \qquad \small{$^{\dag}$corresponding author} \vspace{5pt}\\
\textsuperscript{\rm 1}{FNii, CUHKSZ} \qquad \textsuperscript{\rm 2}{Tencent}
\qquad \textsuperscript{\rm 3}{Tencent-Hunyuan3D}
\qquad \textsuperscript{\rm 4}{SSE, CUHKSZ} \vspace{5pt}\\
\href{https://haolinliu97.github.io/Stable-Score/}{haolinliu97.github.io/Stable-Score}
}
\begin{document}
\twocolumn[{%
\renewcommand\twocolumn[1][]{#1}%
\maketitle
\begin{center}
    \centering
    \captionsetup{type=figure}
    \includegraphics[width=1.0\textwidth]{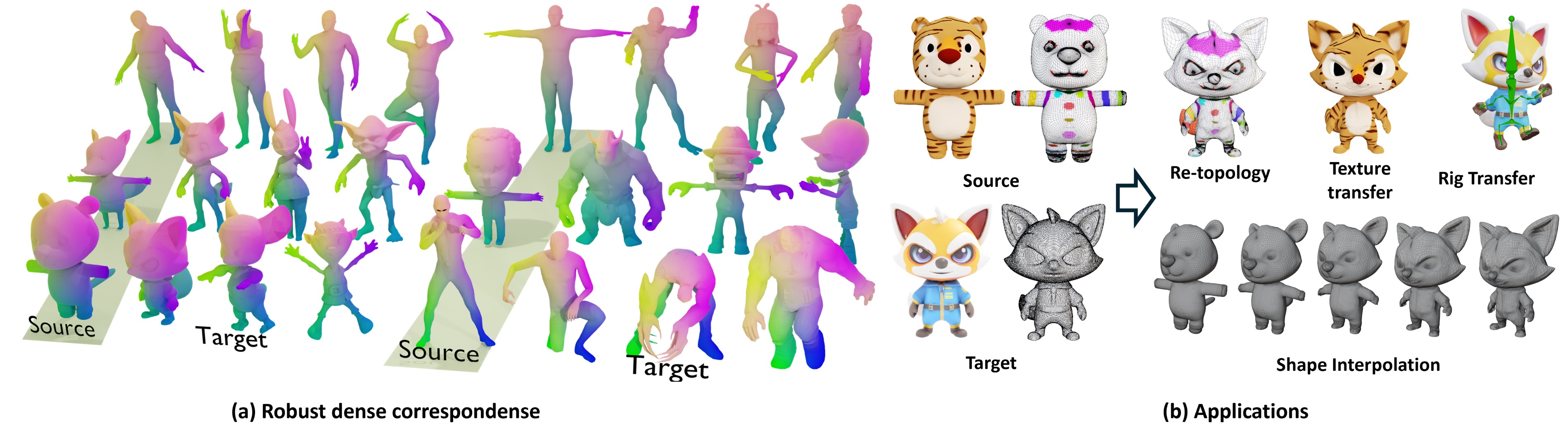}
    \captionof{figure}{
    We propose Stable-SCore: A Stable Registration-based Framework for 3D Shape Correspondence. Given a source and a target mesh, our approach registers the source mesh to the target and establishes dense correspondence between them, with strong robustness to large variations in mesh topology, shape, and pose, as shown in (a). Our method enables a range of downstream applications, including re-topology, texture transfer, rig transfer, and shape interpolation, as shown in (b).
    \label{fig:teaser}}
    \vspace{-0.3cm}
\end{center}%
}]

\begin{abstract}
Establishing character shape correspondence is a critical and fundamental task in computer vision and graphics, with diverse applications including re-topology, attribute transfer, and shape interpolation. Current dominant functional map methods, while effective in controlled scenarios, struggle in real situations with more complex challenges such as non-isometric shape discrepancies. In response, we revisit registration-for-correspondence methods and tap their potential for more stable shape correspondence estimation. To overcome their common issues including unstable deformations and the necessity for careful pre-alignment or high-quality initial 3D correspondences, we introduce Stable-SCore: A Stable Registration-based Framework for 3D Shape Correspondence. We first re-purpose a foundation model for 2D character correspondence that ensures reliable and stable 2D mappings. Crucially, we propose a novel Semantic Flow Guided Registration approach that leverages 2D correspondence to guide mesh deformations. Our framework significantly surpasses existing methods in challenging scenarios, and brings possibilities for a wide array of real applications, as demonstrated in our results. %\xh{2nd round done.}
%There are two main approaches to address this problem: registration-based and functional map-based. While registration-based methods are straightforward and well-suited for handling non-isometric deformations, research in shape correspondence has predominantly focused on similarity-based methods, such as functional maps.
%This trend appears counterintuitive in the era of 3D generative AI, where generated 3D assets frequently exhibit non-isometric shape variations and diverse topologies. 
%In this paper, we propose Stable-Score: A Stable Registration-based Framework for 3D Shape Correspondence.
\end{abstract}    
\section{Introduction}
\label{sec:intro}
% correspondence
% character correspondence 
% challenge, in-the-wild challenge. toy datasets. generatilizaion
Shape correspondence is a fundamental task in computer vision and graphics, where the goal is to establish accurate point-to-point mappings between different shapes, ensuring the preservation of their geometric characteristics. This task involves the identification and alignment of corresponding points, features, or regions across multiple shapes, accommodating variations in pose, scale, or intricate local geometric details. Dense correspondences becomes especially crucial for 3D characters~\cite{sun2023spatially,li2022learning,cao2023unsupervised,cao2024spectral,DeepFunctionalMapsPrior}
, given its implications for a range of downstream applications such as re-topology, shape interpolation~\cite{deng2021deformed,zheng2021deep}, texture transfer~\cite{hormann2007mesh}, rig transfer~\cite{sun2022human,paravati2016point}, \textit{etc}.
%These applications underscore the importance of robust and precise correspondence methods in enhancing the functionality and versatility of 3D character manipulation.

\begin{figure}[tb]
  \centering
  \includegraphics[width=.98 \linewidth]{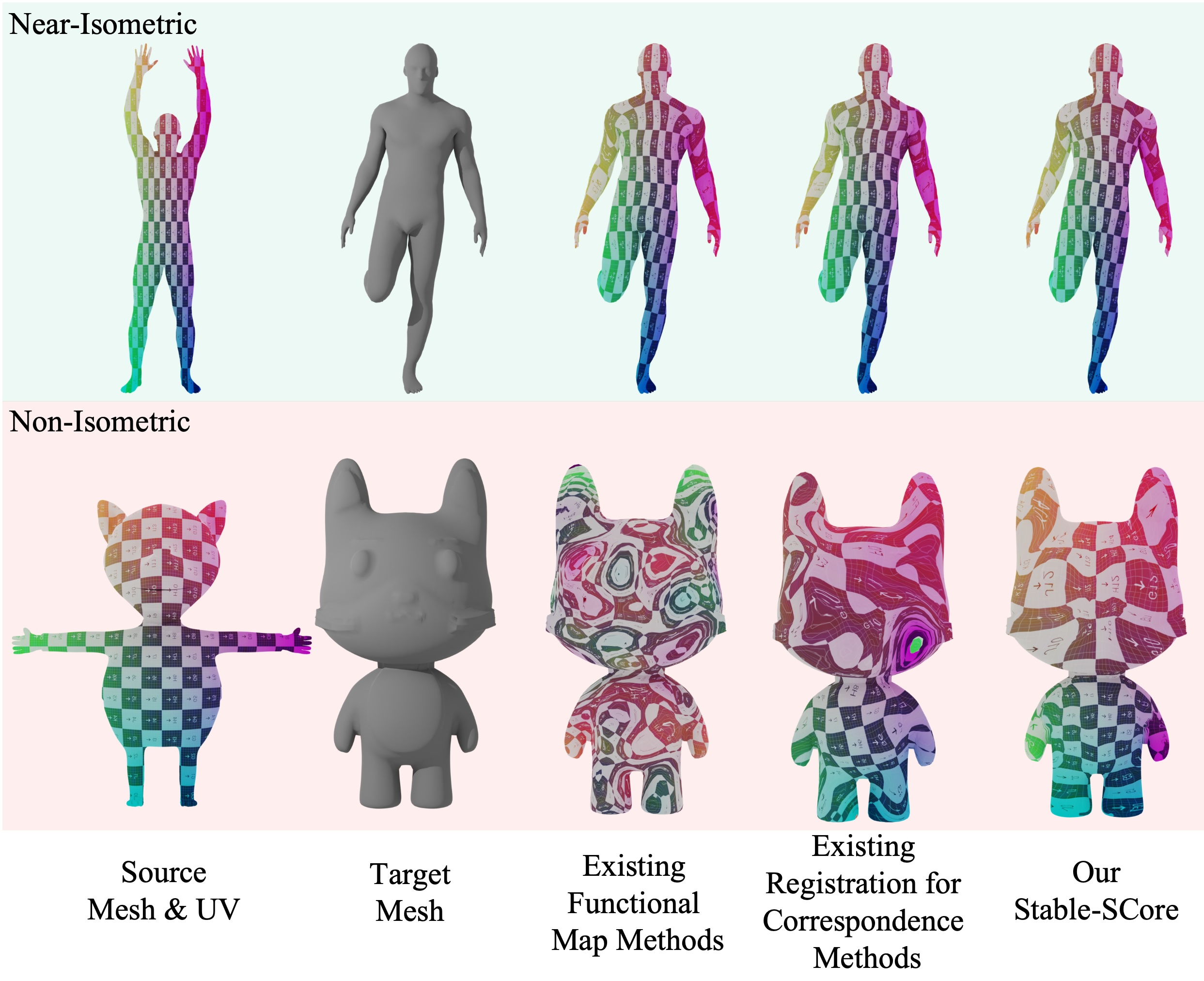}
  \caption{Functional map methods face difficulties with non-isometric correspondences, while existing Registration-for-Correspondence methods break down without reliable initial correspondences. Our Stable-SCore maintains geometric fidelity and offers greater robustness to non-isometric shape variations in the wild.}
  \label{fig:intro}
  \vspace{-0.3cm}
\end{figure}

There are two primary categories to address shape correspondence: registration-for-correspondence methods~\cite{smoothshell,deepshell,registrationfm,bernard2020mina,eisenberger2019divergence,nonrigidicp,li2008global} and functional map methods~\cite{cao2023unsupervised,HybridFunctionalMaps,donati2020deepGeoMaps,li2024deformable,hartwig2023elastic,ovsjanikov2012functional,melzi2019zoomout,roufosse2019unsupervised,DeepGeometricFunctionalMaps}. 
In recent years, functional map methods have dominated this task. These methods transform the challenge of point mapping into one of function mapping, demonstrating leading performance in ``controlled`` scenarios where discrepancies in shape and mesh topology are small compared to those found in 3D models crafted by artists or generated by AI.
Though some of recent work~\cite{cao2023unsupervised,cao2024spectral,HybridFunctionalMaps} claim applicability to non-isometric settings, their performance falls short when being tested on more challenging non-isometric cases, as shown in Figure~\ref{fig:intro}. 
This observation makes us rethink the suitability of functional maps for such tasks.
Also highlighted in recent studies~\cite{cao2023unsupervised,bernard2020mina}, functional map methods encounter difficulties with non-isometric correspondence due to their reliance on strictly aligned low-rank basis.

In contrast, the registration-for-correspondence paradigm is essentially more adept at handling non-isometric discrepancies. Given high-quality initial correspondences, these methods~\cite{nonrigidicp,deformpyramid} iteratively deform the source shape to align it with the target shape, ultimately producing dense correspondences. Despite their straightforward and intuitive approach, registration-for-correspondence methods encounter several challenges. First, existing mesh deformation techniques~\cite{nonrigidicp,arap,smoothshell,eisenberger2019divergence} often result in unstable transformations, suffering from distortion artifacts and frequently struggling to strike a balance between smoothness and deformation accuracy. Secondly, these methods typically depend on careful pre-alignment, or high-quality initial sparse 3D correspondences, which are challenging to secure, particularly when the shapes undergo substantial non-isometric deformations, as depicted in Figure~\ref{fig:intro}. These limitations have hindered the broader adoption and effectiveness of the registration-for-correspondence paradigm.
%Second, it limits the application since it often requires rough alignment or high-quality initial correspondences, which are difficult to obtain.

%In this work, we rethink the registration approach to address these challenges, with the goal of achieving a stable registration process. Our approach begins by employing a stable mesh deformation model. 
Embracing registration-for-correspondence as a foundational paradigm, this work reevaluates and addresses its inherent challenges.
%
%It essentially comprises two basic elements, stable deformation mechanism and reliable guidance signals.
(1) Initially, we integrate the emerging deformation technique known as Neural Jacobian Fields (NJF)~\cite{njf} as our deformation model. NJF is recognized for producing stable deformations and is well-suited for applications involving both differentiable rendering and iterative optimization~\cite{gao2023textdeformer,wan2022meshup,wang2025headevolver}.
(2) Given the scarcity of large-scale 3D correspondence datasets, robust initial correspondence estimation seems impractical. However, recent breakthroughs in re-purposing 2D foundational models have demonstrated remarkable capabilities across various 2D tasks such as depth estimation~\cite{marigold,fu2025geowizard,genpercept} and novel view synthesis~\cite{zero123, shi2023zero123++, wang2023imagedream, voleti2025sv3d, zuo2024videomv, liu2024reconx}. Leveraging this advancement, we train a 2D character correspondence model designed to establish stable 2D correspondences.
(3) Crucially, we introduce a novel Semantic Flow Guided Registration framework, equipped with the aforementioned deformation model and the 2D character correspondence model. Within this framework, deformations are rendered in a differentiable manner and supervised using 2D correspondences. Via iterative optimization, the source mesh is progressively deformed to align the shape of the target mesh.
In this way, this framework produces stable 3D dense correspondence under challenging non-isometric settings.

Furthermore, to enhance robust in-the-wild character correspondence - a practical need in real-world applications - we introduce a challenging new benchmark, \textit{Character in-the-wild~(CharW)}. This benchmark consists of meshes from various sources, including professional artists and text-to-3D generation approaches~\cite{li2025triposg,zhang2024clay}. It features significant geometric and topological variations, capturing a broad spectrum of non-isometric deformations. This diversity ensures that \textit{CharW} effectively simulates the complex scenarios encountered in practical settings, making it essential for advancing the field of character correspondence. %Additionally, most meshes include textures and rigging information, facilitating downstream applications such as character animation transfer and texture transfer.

In summary, our key contributions are as follows:
\begin{itemize}
% framwork: 1) 2d correpondence foundation model for character, strong generalization 2) deformation module
% \item We propose **XXX**, a novel framework that leverages both 2D foundation model priors and extensive training on dense correspondence datasets, demonstrating robust in the wild generalization capabilities.
%\item We make registration great again for 3D character correspondence by introducing a 2D correspondence foundation model for the character, a strongly regularized deformation module, and a seamless supervision method that effectively bridges correspondence with deformation.
\item We revisit the registration-for-correspondence paradigm, and propose a novel framework, Stable-SCore, to address its key challenges and revitalize its effectiveness.
% (放到正文) benchmark: 1) more practical problem, we propose a more challenge benchmark, 
% \item We introduce a new shape correspondence benchmark featuring shapes with significant variation in geometry and topology, including those generated by state-of-the-art 3D generation techniques. We hope this could facilitate research into more robust shape correspondence methods.
\item We introduce \textit{Character in-the-wild~(CharW)}, the first benchmark for character correspondence in-the-wild, featuring shapes with significant geometric and topological variations.
% SOTA on all benchmarks, expecially on more challenge benchmarks.
\item Through comprehensive experiments, Stable-SCore shows state-of-the-art performance across all non-isometric character correspondence benchmarks, significantly surpassing previous methods.
\item We highlight several downstream applications powered by our approach, such as re-topology, texture transfer, rig transfer, and shape interpolation. These applications showcase the potential of Stable-SCore to open new avenues for a variety of creative and practical uses.
\end{itemize}

\section{Related Work}

\subsection{2D Image Correspondence}
2D dense correspondence refers to the problem of establishing pixel-wise correspondences between two or more images, based on visual or geometric similarity. Traditional methods often relied on handcrafted feature descriptors such as SIFT~\cite{sift} or SURF~\cite{bay2006surf}. Recent advancements in deep learning have revolutionized by learning feature representations directly from data~\cite{cho2021cats,huang2022learning,kim2022transformatcher,lee2021patchmatch,jabri2020space,wang2020learning,wang2019learning,tumanyan2022splicing}. However, due to the limited capacity of the feature extractor and data, their methods struggle with domain gap problem whey applied to in-the-wild images.

Recently, 2D visual foundation models like DINO~\cite{oquab2023dinov2} and Stable Diffusion~\cite{rombach2022high,stablediffusion} have demonstrated remarkable generalization capabilities in the image dense correspondence task, outperforming previous 2D correspondence methods~\cite{amir2021deep,gupta2023asic,hedlin2024unsupervised,luo2024diffusion,tang2023emergent,zhang2024tale,zhang2023telling}. Inspired by~\cite{zhang2024tale,zhang2023telling}, we employ a lightweight adapter network that leverages features from Stable Diffusion and DINO to train a 2D character correspondence model.

% correspondence and registration (NJF)
% 3D correspondence and registration
\subsection{3D Shape Correspondence and Registration}
% Shape correspondence refers to the problem of establishing a meaningful mapping between two or more shapes, such that each point on one shape is matched with a corresponding point on another shape. Existing methods can be broadly categorized into two main groups: (1) axiomatic methods and (2) learning-based methods.
Shape correspondence methods can be broadly categorized into two main approaches: functional map methods and registration methods. 
The key idea behind functional maps~\cite{ovsjanikov2012functional} is to express correspondences not as point-to-point matches, but as mappings between functions. FMNet~\cite{DeepGeometricFunctionalMaps} is the first to combine deep learning and functional map. DiffusionNet~\cite{sharp2022diffusionnet} proposes an effective spectral feature extractor that pushes the deep functional map methods further. Follow-up works such as~\cite{DeepFunctionalMapsPrior,DeepGeometricFunctionalMaps,sun2023spatially,li2024deformable,attaiki2023shape,attaiki2023understanding,magnet2023scalable} extend it and enhance the performance. Some studies~\cite{hartwig2023elastic,HybridFunctionalMaps,cao2023unsupervised,DeepFunctionalMapsPrior,cao2024spectral,eisenberger2021neuromorph,smoothshell,deepshell} attempt to address non-isometric correspondence by considering different basis functions or involves extrinsic information. However, they still struggle with the non-isometric setting.

Many axiomatic correspondence techniques rely heavily on registration. Among these, the Non-Rigid ICP~\cite{nonrigidicp,li2008global} are widely used due to their effectiveness in non-rigid registration tasks. \cite{deformpyramid} propose a point-based hierarchical deformation field for registration. \cite{eisenberger2019divergence} propose a divergence-free deformation field, and alternatively updates correspondence and registration. \cite{bernard2020mina} find correspondences and registers based on control points. \cite{smoothshell} proposed a low-rank deformation combining functional map. Recently, some registration-based methods have sought the power of neural networks. \cite{deepshell} uses spectral convolution filters to process mesh data, and the output features provide initial correspondence for the registration process as in \cite{smoothshell}. \cite{DeepFunctionalMapsPrior} use functional map prior for registration through a deformation graph. However, these consistently face some challenges such as unstable deformation processes and require rough alignment or initial correspondence, which limits its application in shape correspondence. 
%\textcolor{red}{To address these problems, we propose Stable-SCore: A Stable Registration-based Framework for 3D shape Correspondence. Specifically, we advocate a recent advanced deformation framework Neural Jacobian Fields~\cite{njf}, and propose to train a 2D character correspondence model to guide the deformation.}\xh{I think the read part can be deleted.}

There are some other methods~\cite{roetzer2024spidermatch,roetzer2022scalable,roetzer2023fast,windheuser2011geometrically,ehm2024geometrically} who attempt to solve a geometric consistent shape correspondence, in which connection remains after the mapping. \cite{roetzer2024spidermatch} propose a 3D-2D mapping method and solve it by finding the shortest path in the product graph. This kind of method shows potential in non-isometric correspondence, however, they are only suitable for low-poly meshes.

\subsection{Repurposing 2D Models for 3D Tasks}
Recently, 2D vision foundation models have emerged, driven by the availability of billion-scale training datasets. Notable examples include CLIP~\cite{clip}, DINOv2~\cite{oquab2023dinov2}, Stable Diffusion~\cite{stablediffusion}, and SAM~\cite{kirillov2023sam}. In contrast, obtaining datasets of similar scale for building foundation models in 3D vision remains a significant challenge. As a result, many recent works have sought to leverage the powerful generalization capabilities of these 2D foundation models to address 3D tasks, such as depth estimation~\cite{marigold,fu2025geowizard,genpercept} and novel view synthesis~\cite{zero123, shi2023zero123++, wang2023imagedream, voleti2025sv3d, zuo2024videomv, liu2024reconx}. 
For instance, Marigold~\cite{marigold} proposes fine-tuning the U-Net of Stable Diffusion for depth estimation by concatenating the input image latent with a noisy latent representation. \cite{sun2024srif} propose to use Diffmorpher\cite{zhang2024diffmorpher} to distill 2D morphing to 3D morphing for shape correspondence.
A closely related work, Diff3f~\cite{Diff3f}, back-projects 2D foundational features onto 3D surfaces, and utilize these 3D features for shape correspondences. However, it still requires using the functional map technique to obtain a continuous map. This approach results in a loss of semantic information, as the features are projected onto a low-rank basis, leading to degradation in detail.
In contrast, our method re-purposes the foundation model for dense 2D correspondences, with a sophisticated design to guide registration through differentiable rendering, preserving as much detail as possible.
% repurpose 2D for 3D task
\section{Method}
\begin{figure*}[ht]
\includegraphics[width=1.0\textwidth]{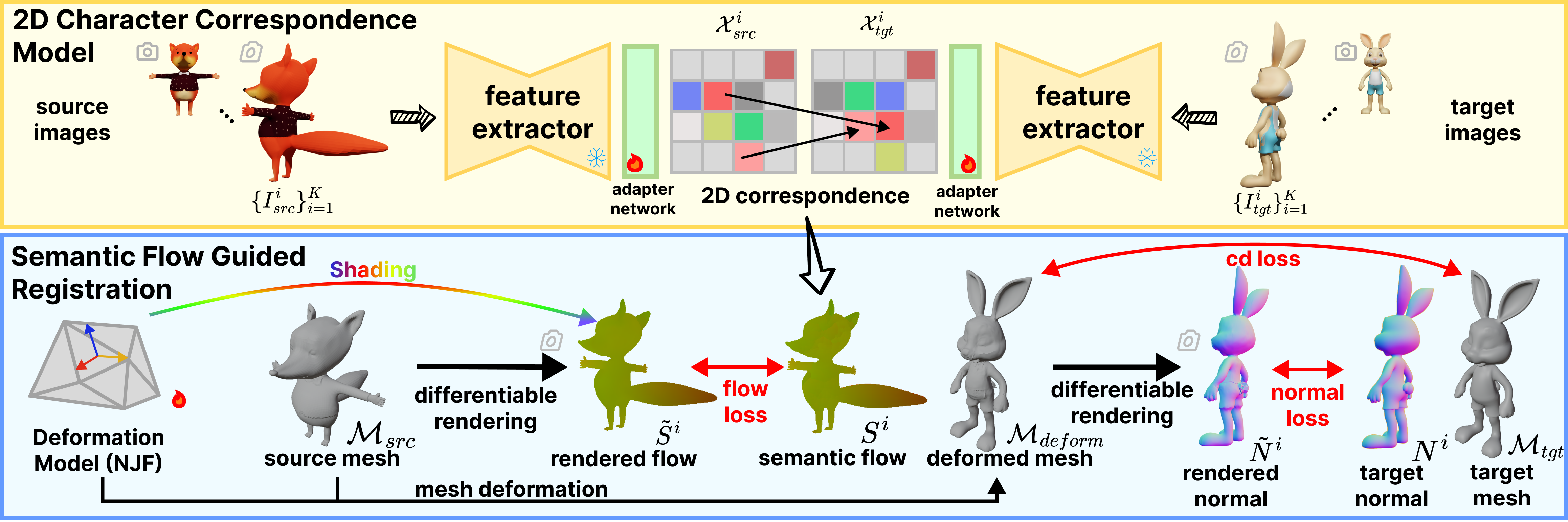}
\captionof{figure}{
    The Stable-SCore pipeline operates as follows: Source and target meshes are inputted and rendered into multi-view RGB or normal images using a fixed set of cameras. These images are processed through the network to extract 2D correspondences as a semantic flow map. 
     Using differentiable rendering, we render forward flow under the same camera views and supervise it with the semantic flow. Chamfer Distance (CD) and normal loss, are also applied. The deformation model is iteratively optimized throughout this process.
    \label{fig:method}}
\end{figure*}
\subsection{Overview}
% We propose Stable-Score: A Stable Registration-based Framework for 3D Shape Correspondence. Given a source mesh $\mathcal{M}_{src}$ and a target mesh $\mathcal{M}_{tgt}$ as inputs. Our method first re-purposes 2D foundation models (Sec. \ref{sec:2dcorr}), to extract robust multi-view 2D correspondence. Next, we introduce a Mesh Deformation Module (Sec. \ref{sec:deform}) to drive the registration process. Finally, we employed a Semantic Flow guided Registration method (Sec. \ref{sec:register}), to guide the registration process using 2D correspondence.
We propose Stable-SCore: A Stable Registration-based Framework for 3D Shape Correspondence. Given a source mesh $\mathcal{M}_{src}$ and a target mesh $\mathcal{M}_{tgt}$ as inputs, our method first re-purposes 2D foundation models (Sec. \ref{sec:2dcorr}) to extract robust multi-view 2D correspondences. Then, we employ a Semantic Flow-guided Registration method (Sec. \ref{sec:register}) to guide the registration using the 2D correspondences.

\subsection{2D Character Correspondence Model\label{sec:2dcorr}}
% In order to obtain accurate and robust 2D correspondence, we first re-purpose 2D foundation models Stable Diffusion~\cite{stablediffusion} and DINO~\cite{oquab2023dinov2}. These two foundation models serve as the feature extractor as in Figure~\ref{fig:method}. 
% We assume the input meshes are aligned in rotation or adjusted using estimated rotation as in~\cite{zhang2023telling} so that both target and source meshes are aligned under the same view. 
To achieve accurate and robust 2D correspondences, we adopt the pre-trained foundation models, specifically Stable Diffusion~\cite{stablediffusion} and DINO~\cite{oquab2023dinov2}, for correspondence estimation. These models act as feature extractors, followed by an adapter network to extract 2D correspondences, as illustrated in Figure~\ref{fig:method}.

In detail, for given source and target meshes that are roughly aligned in rotation, we establish a fixed set of camera poses $C_i, i=1,2,...,K$, Then, we render multi-view images ${(I_{src}^i,I_{tgt}^i),i=1,2,...,K}$ for both source and target as input images. The input images can be either rendered as RGB or normal images. We feed these images into Stable Diffusion and DINO to extract feature maps. Specifically, DINO features are collected from the output of DINO, while Stable Diffusion features are extracted from the intermediate layers of the UNet.

Subsequently, a lightweight adapter network is employed to map the source and target features into a common embedding space. Within this space, 2D correspondences are determined through a nearest neighbor search. These correspondences are then encapsulated in a semantic flow map, which illustrates the 2D displacements linking source pixels to their respective target pixels.

In this way, for each camera pose $C_i$, we find 2D correspondences between $I_{src}^i$ and $I_{tgt}^i$, producing a semantic flow map $S^i$.

\paragraph{Training the 2D correspondence model.} 
% We train the adapter network while fixing the feature extractors. The network is trained not only on 3D character correspondence dataset 3DBiCar~\cite{luo2023rabit} and Surreal~\cite{varol17_surreal} but also on 2D correspondence dataset SPair 71K~\cite{min2019spair}. 3DBiCar and Surreal datasets contain parametric models sharing the same topology, which enables retrieving 3d correspondence between each pair of samples. The 3d correspondences are projected into the image plane as 2D correspondences for supervision. By utilizing various training data, it is able to achieve robust and accurate 2D correspondence for characters. During the training, we use CLIP loss as in \cite{zhang2023telling}. However, using only the CLIP loss will cause self-similarity problems (left hand is similar to right hand) as it treats all negative samples equally. To address this problem, we propose a geometry-grounded negative loss. For each source vertices, we take another vertex that is far away to be hard negative samples. Geodesic distance is used as the distance metric. 
We train the adapter network while keeping the feature extractors fixed. The network is trainable using both 3D character correspondence and 2D correspondence datasets, including 3DBiCar~\cite{luo2023rabit}, Surreal~\cite{varol17_surreal}, and SPair-71K~\cite{min2019spair}. We believe that with diverse training data, the network is able to estimate robust and accurate 2D correspondences for characters.

During training on 3D character correspondence datasets, we first randomly sample a pair of meshes as the source and the target, and render a pair of images $(I_{src}, I_{tgt})$ given random camera poses $(C_s, C_t)$ sampled from a prior camera distribution. We force $C_s=C_t$ with a 50\% probability. We randomly render normal or RGB images as inputs to train the model. The ground-truth 3D correspondences are projected onto the image plane using $(C_s, C_t)$ to create matched 2D points for supervision.
For 2D character correspondence, the pairs of $(I_{src}, I_{tgt})$ are sampled from the dataset. 

$(I_{src}, I_{tgt})$ are fed into the feature extractor and the adapter network to produce the embeddings $(\mathcal{X}_{src}, \mathcal{X}_{tgt})$.
During training, we use a contrastive loss as described in \cite{zhang2023telling}:
\begin{equation}
\mathcal{L}_{con}=CL(\mathcal{X}_{src}(\mathcal{P}^s),\mathcal{X}_{tgt}(\mathcal{P}^t))
\end{equation}
where $CL$ is the CLIP-style contrastive loss, $\mathcal{P}^s$ and $\mathcal{P}^t$ are pairs of matched 2D points, $\mathcal{X}_{src}(\cdot)$ and $\mathcal{X}_{tgt}(\cdot)$ denote the processes to sample point features from $\mathcal{X}_{src}$ and $\mathcal{X}_{tgt}$.

However, relying solely on contrastive loss introduces self-similarity issues, \eg, the left hand being treated similarly to the right hand: different parts may be semantically similar. 
To address this, we propose a geometry-grounded negative loss. The idea is to reduce self-similarity for a pair of points that are geodesically far away on two meshes respectively.  %For each source vertex, we try to sample the other vertex on the target mesh that is geodesically far away as a negative sample. However, it is impossible to measure the geodesic distance of two vertices from different meshes. 
For each pair of points, we are able to compute the pseudo geodesic distance on the common mesh template since we use parametric models such as RaBit~\cite{luo2023rabit} and SMPL~\cite{SMPL:2015} for training.
This loss can be described by the following equation:
\begin{equation}
\mathcal{L}_{neg}=\sum_{(p,q),\mathbb{G}(p,q)>th}\|\mathcal{X}_{src}(\Pi(p,C_s))\cdot\mathcal{X}_{tgt}(\Pi(q,C_t))\|_2
\end{equation}
where $\mathbb{G}$ is the pseudo geodesic distance matrix pre-computed on the common mesh template, $th$ is the threshold distance, $(p,q)$ are randomly sampled vertices from the source mesh and target mesh respectively, $\Pi(\cdot)$ denotes the projection operator. The geometry-grounded negative loss is able to alleviate the self-similarity problem and finally achieve more robust shape correspondence. Thus, the total loss to train the 2D character correspondence model is defined as $L_{2D} = L_{con} + \lambda_{neg}L_{neg}$ where $\lambda_{neg}$ is a weighting factor.

% \paragraph{Leveraging 2D correspondence model.}
% xxx

\subsection{Semantic Flow Guided Registration\label{sec:register}}
% A naive approach to utilize the 2D foundation model for 3D correspondence involves back-projecting the 2D foundational features onto the 3D surface, performing multi-view averaging, and then applying functional map techniques to obtain the final 3D continuous correspondence, as demonstrated in \cite{Diff3f}.
% However, this approach faces several challenges. As the multi-view averaging and low-rank basis projection in the functional map techniques destroy the integrity of the features. \Haolin{Is this statement too heavy?}\xh{remove}

% First, multi-view averaging operation harms features' integrity due to multi-view inconsistency problem as stated in \cite{} \Haolin{To zizheng: help me add some citations in here for the multi-view inconsistency problem}. 
%First, the multi-view averaging operation tends to over-smooth the features, which can lead to a loss of detail. Second, the raw foundation model features produce noisy 2D correspondences, which persist even after being lifted to 3D. Third, the method still requires the functional map to obtain a continuous correspondence, but projecting the features onto a low-rank basis further compromises their integrity.\xh{Too verbose. Reduce to 2-3 sentences.}

\paragraph{Mesh Deformation Model.\label{sec:njf}}
Registration-based methods often involve iteratively deforming a source mesh to align it with a target mesh. The mesh deformation model is essential for successful registration, as it aims to produce a smooth deformed result that preserves details while minimizing distortion. We choose Neural Jacobian Fields (NJF)~\cite{njf} as our deformation model to drive the registration, as it provides powerful and stable deformations, as demonstrated in \cite{wang2025headevolver,wan2022meshup,gao2023textdeformer}.
Now we introduce some preliminaries of NJF. It first defines per-face Jacobian $J_i$ as follows:
\begin{equation}
J_i\mathcal{B}_i^T [v_k-v_j,v_l-v_j] = [\phi_k-\phi_j,\phi_l-\phi_j]
\label{eq:jacobian}
\end{equation}
where $v_k$, $v_j$, $v_l$ are the original vertices' location of a triangle, $\phi_k$, $\phi_j$, $\phi_l$ are the deformed vertices' location, $J_i \in \mathbb{R}^{3 \times 2}$ is the face Jacobian, $\mathcal{B}_i \in \mathbb{R}^{3 \times 2}$ are two-column vectors that form an orthonormal basis for the tangent space of the face. This equation further defines a gradient operator as follows:
\begin{equation}
J_i=\Phi\nabla_i^T,
\end{equation}
where $\Phi \in \mathbb{R}^{n\times3}$ are vertices deformed locations. Solving the following Poisson equation yields the deformed vertex locations:
\begin{equation}
\Phi^\ast = \argmin_\Phi\sum_i |t_i|\|\Phi\nabla_i^T-J_i\|_2
\end{equation}
Where $|t_i|$ is the face's area. This Poisson equation can be solved in a least-square approach by the following equation:
\begin{equation}
\Phi^\ast=L^{-1}\mathcal{A}\nabla^T J
\label{eq:least_square}
\end{equation}
where $L$ is the mesh's cotangent Laplacian, $A$ is the mesh's mass matrix and $J$ is the stack of estimated Jacobian $J_i$. During the registration process, we optimize the per-face transformation matrices $\tilde{J}_i \in \mathbb{R}^{3 \times 3}$. These matrices are then projected onto each face's tangent space $\mathcal{B}_i$ to obtain the Jacobian $J_i$ by $J_i=\tilde{J}_i\mathcal{B}_i$. Finally, it applies Equation~\ref{eq:least_square} to solve the final vertices location $\Phi^\ast$.
% The intuition of the Neural Jacobian Field is to define the deformation in a compact parameterized tangent space, which makes it easier for optimization compared with it in ambient space. Please refer to \cite{njf} for more details.
The intuition behind NJF is to define the deformation in a compact, parameterized tangent space, which simplifies the optimization process compared to performing it in the ambient space. For more details, please refer to \cite{njf}.

% Now, the gradient operator $\nabla_i$ can be derived as:
% \begin{equation}
% J_i=\Phi\nabla_i^T
% \end{equation}
% Now, given a prediction or estimated Jacobian matrices $\tilde{J}_i$ of each face, we can solve the below Poisson equation to obtain the optimized deformed location $\Phi^\ast$:
% \begin{equation}
% \Phi^\ast = \argmin_\Phi\sum |t_i|\|\Phi\nabla_i^T-M_i\|^2
% \end{equation}
% This Poisson equation can be solved in a least square manner by the following equation:
% \begin{equation}
% \Phi^\ast=L^{-1}\mathcal{A}\nabla^T \tilde{J}
% \end{equation}
% $L$ is the mesh's cotangent Laplacian. $A$ is the mesh's mass matrix. $\tilde{J}$ is the stack of predicted or estimated jacobian $\tilde{J}_i$. In practice, it will optimize a $\tilde{M}_i \in \mathbb{R}^{3 \times 3}$ per-face transformation matrices and further restrict them to the tangent space of the triangle, to form the $\tilde{J}_i$ used to solve the Poisson equation. We will abuse the notation $\tilde{J}_i$ as the $3 \times 3$ transformation matrices before being restricted in the following section for brevity. 

\vspace{-2mm}
\paragraph{Semantic flow guidance.} 
%\textcolor{red}{We aim to design a framework that preserves integral information to achieve high-quality and continuous 3D correspondence. Therefore, we propose Semantic Flow Guided Registration. The core idea is to use 2D correspondence to supervise the deformation processes.}\xh{Red part deleted.} 
%As aforementioned, we choose Neural Jacobian Field~\cite{njf} as our deformation model and opt for iterative optimization of per-face transformation matrices $\tilde{J}_i$. These transformation matrices can be converted to deformed vertices location $\Phi^\ast$ as in Equation~\ref{eq:least_square}. 

Given the deformed vertices $\Phi^\ast$ from NJF, we project the vertices and compute the displacement to shade the source mesh $\mathcal{M}_{src}$ and utilize NVDiff~\cite{Laine2020diffrast} to render a differentiable flow map $\tilde{S}^i$. The shading and rendering procedure can be described as follows:
\begin{align}
F_i &=\Pi(\Phi^\ast,C_i)-\Pi(\mathcal{V}_{src},C_i) \label{2ddeform} \\
\tilde{S}^i &= R(\mathcal{M}_{src}, F_i, C_i)
\end{align}
where $\Phi^\ast$ is the deformed vertices location while $\mathcal{V}_{src}$ are source vertices location, $C_i$ is the camera pose same as that used in 2D character correspondence model. We first compute the per-vertex 2D displacement $F_i$ by projecting vertex coordinates into 2D coordinates using the projection operator $\Pi$ and subtracting them by Equation~\ref{2ddeform}. We further use vertex color shading and assigned the normalized 2D displacement as the vertex color. Finally, a differentiable rendering function $R(\cdot)$ is used to render the flow map $\tilde{S}_i$ under camera pose $C_i$.
The flow loss computed between the multi-view rendered flows and semantic flows $S_i$ yielded from the 2D correspondence model is formulated as follows: 
%\xh{better have formulation for $\tilde{S}^i$, such as $\tilde{S}^i=NVDiff(shade(M_{src}, D), CAM^i), D=SomeFunction({\tilde{J}})$, where $CAM^i$ is the $i$-th camera, $D$ is vertices displacements, $\tilde{J}$ is the estimated Jacobian, so as to make it a complete differentiable pipeline, also to link section 3.1.}
\begin{equation}
\mathcal{L}_{flow}=\sum_{i=1}^{K}\|\tilde{S}^i-S^i\|_1
\end{equation}

\begin{table*}[ht]
    \centering
    \caption{Quantitative comparison between our methods and previous methods. The evaluation metric is mean geodesic error $\times 100$. Ours (Normal) indicates our method with rendered normal images as inputs, aligned with baseline methods that only use geometry, while Ours (RGB) uses rendered RGB images. Ours (Zero-shot) is a variant with adapter network removed and no training is needed. \textbf{Bold} indicates the best method while \underline{underline} indicates the second best.}
    \resizebox{0.85\textwidth}{!}{
    \begin{tabular}{c|l|c|c|c|c|c|c|c}
        \toprule
        \multicolumn{2}{c|}{} & \multicolumn{4}{|c|}{Cross Domain} & \multicolumn{3}{|c}{Intra Domain} \\
        \midrule
        \multicolumn{2}{c|}{} & Isometric & \multicolumn{3}{|c|}{Non-isometric}& Isometric& \multicolumn{2}{|c}{Non-isometric} \\
        \midrule
        \textbf{Supervision} &\textbf{Test dataset} & FAUST & CharW & DT4D-H std &DT4D-H hard& FAUST & DT4D-H std&DT4D-H hard\\
        \midrule
        \multirow{3}{*}{Zero-shot} &SmoothShell~\cite{smoothshell} & 2.93  & 11.6  & \underline{13.6} &\underline{12.4} & 2.93 & 13.6 & 12.4    \\
        &Diff3f~\cite{Diff3f} & 12.0  & 12.5 & 24.0 & 22.7  & 12.0 & 24.0 & 22.7  \\
        &Ours (Zero-shot)  & 5.60  & 3.48 & 19.9 & 14.1  & 5.60 & 19.9 & 14.1  \\
        \midrule
        \multirow{5}{*}{Unsupervised}&DeepShell~\cite{deepshell} & 6.50 & 37.5 & 31.0 & 40.8 & 1.90 & 29.1 & 37.7 \\
        &DFR\cite{DeepFunctionalMapsPrior} & 18.2 & \underline{6.52} & 19.8 & 14.3 & 9.81 & 14.9 & 7.67 \\
        &HybridGeoFMap~\cite{HybridFunctionalMaps} & 6.61  & 32.2 & 22.1 & 29.0 & 2.39 & 4.08 & 4.13   \\
        &ULRSSM~\cite{cao2023unsupervised} & 2.09 & 32.6 & 28.2 & 32.0 & 1.69 & 4.61 & 6.91 \\
        &Hybrid ULRSSM~\cite{HybridFunctionalMaps} & \textbf{1.55}  & 33.5 & 15.5 & 22.1 & \textbf{1.48} & \underline{3.47} & \underline{3.95}  \\
        \midrule
        \multirow{3}{*}{Supervised}&GeoFMap~\cite{donati2020deepGeoMaps}  & 2.81  & 30.2  & 25.2 & 24.5 & 2.65 & 4.12 & 4.21   \\
        &Ours (Normal) & \underline{1.83}  & \textbf{2.61} & \textbf{4.23} & \textbf{4.12} & \underline{1.58} & \textbf{3.11} & \textbf{3.38}    \\
        &Ours (RGB) & - & 2.57 & - & - & -& - & -\\
        \bottomrule
    \end{tabular}}
    \label{tab_compare}
\end{table*}

\vspace{-2mm}
\paragraph{Geometry alignment loss.} 
% Registration loss is designed to produce deformed results that approach the target mesh. We deform the mesh using the vertex displacements and obtain a deformed mesh $\mathcal{M}_{deformed}$. We further compute the Chamfer Distance loss $\mathcal{L}_{cd}$ between vertices of $\mathcal{M}_{deformed}$ and $\mathcal{M}_{target}$. Moreover, we use differentiable rendering to render the normal of $\mathcal{M}_{deformed}$ and compute a normal loss $\mathcal{L}_{normal}$ with the normal rendering of the target mesh.
The geometry alignment loss aims at spatially aligning the shape of the deformed source mesh and the target mesh. We apply vertex displacements to obtain the deformed source mesh, $\mathcal{M}_{\text{deform}}$, and compute the Chamfer Distance loss, $\mathcal{L}_{\text{cd}}$, between its vertices and those of the target mesh, $\mathcal{M}_{\text{target}}$. Additionally, we use differentiable rendering to render the normal map of $\mathcal{M}_{\text{deform}}$ and compute the normal loss.

\vspace{-2mm}
\paragraph{Deformation Regularization.} 
To stabilize deformation, we use an identity-preserving term enforced on the per-face transformation matrix as in \cite{gao2023textdeformer}:
\begin{equation}
\mathcal{L}_{identity}=\sum_i^{|\mathcal{F}|}\|\tilde{J}_i-I_3\|_F
\end{equation}
% However, only using this identity-preserving term is not enough. It will hinder large pose deformation if we apply a large weighting factor. On the other hand, the deformation is not smooth if a small weight is applied. 
However, using only this identity-preserving term is not sufficient. Applying a large weighting factor will impede significant pose deformation while applying a small weight results in unsmooth deformation.
To address this problem, we propose a shear-resistant term to regularize the deformation, as described below:
% \Haolin{Is there any better names?}
\begin{equation}
\mathcal{L}_{shear}=\sum_i^{|\mathcal{F}|}\|\tilde{J}_i-\tilde{J}_i^{rot}\|_F,
\end{equation}
where $\tilde{J}_i^{rot}$, the rotational component of $\tilde{J}_i$, is extracted via polar decomposition. This is inspired by the observation that nice deformation results are majorly contributed by rigid transformation and avoid shear deformation.

\paragraph{Optimization.}
The final loss is formulated as follows:
\begin{equation}
\begin{aligned}
\mathcal{L}=\lambda_{flow}\mathcal{L}_{flow} + \lambda_{cd}\mathcal{L}_{cd}+\lambda_{normal}\mathcal{L}_{normal} \\
+\lambda_{identity}\mathcal{L}_{identity} + \lambda_{shear}\mathcal{L}_{shear}
\end{aligned}
\end{equation}
%\xh{Incomplete. How is the whole system optimized? How is the final 3D correspondence result derived from optimized NJF? Use 2-3 sentences to describe these.}
We iteratively optimize the per face transformation matrix $\tilde{J}_i$ via minimizing the above loss. After the optimization converges, We use the procedure described in Equation~\ref{eq:least_square} to obtain the registration results $\Phi_{final}$, To retrieve 3d correspondences, we first find the nearest face in $\mathcal{M}_{tgt}$ for each $\Phi_{final}^i$. Next, $\Phi_{final}^i$ is projected onto the nearest face and the barycentric coordinate is computed to establish the final mapping between $\mathcal{M}_{src}$ and $\mathcal{M}_{tgt}$.
%Our method significantly outperforms \cite{Diff3f}, as demonstrated in the experiment in Sec.~\ref{sec:compare}. As our method ensures information integrity through 2D correspondence supervision, while smooth deformation produce high-quality continuous mapping. \Haolin{check if this ok}\xh{remove}
%The reasons are threefold: First, the smooth deformation produces continuous correspondences. Second, this smooth deformation adapts better to noisy correspondences. Third, we design a semantic flow guidance that directly utilizes raw 2D correspondences as supervision, ensuring the integrity of the information.\xh{Too verbose, remove or reduce it.}

% preliminary

% 2D foundation correspondence

% Mesh deformation
%\input{sec/benchmark}
\section{Experiments}
\subsection{Implementation Details} %\xh{can move to supp if pages not enough.}

To train the 2D character correspondence model, we first render 30 views each for the source and target meshes. Azimuth angles are sampled at 60-degree intervals, and elevation angles range from -30° to 50°. The resulting images are passed through DINOv2~\cite{oquab2023dinov2} to extract $60 \times 60$ feature maps. These images are also encoded using the VAE encoder from Stable Diffusion 1.5~\cite{stablediffusion}. After adding noise at timestep $t=50$, the latent images are processed by the U-Net, from which features are extracted from the upsampling layers. Using a sliding window strategy, we obtain $120 \times 120$ features from the last upsampling layer. Features from earlier upsampling layers and DINOv2 are upsampled and concatenated, forming the final $120 \times 120$ feature map input to the adapter network. The weights for the geometry-grounded negative loss are set to $\lambda_{neg} = 5.0$. During inference, the adapter’s output is upsampled to $512 \times 512$ and used to compute the semantic flow map.

In the Semantic Flow Guided Registration stage, we optimize the Neural Jacobian Fields for 5,000 iterations, which takes approximately 2 minutes for meshes with 10K faces and 4 minutes for those with 40K faces. We use differentiable rendering to generate $512 \times 512$ flow and normal maps, which are used to compute flow and normal losses. The loss weights are: $\lambda_{flow} = 10.0$, $\lambda_{cd} = 1.0$, $\lambda_{normal} = 0.1$, and $\lambda_{shear} = 0.1$. The identity-preserving loss weight $\lambda_{identity}$ decays linearly from 0.01 to 0.0001 throughout the optimization.

\begin{figure*}[htb]
  \centering
  \includegraphics[width=.99 \linewidth]{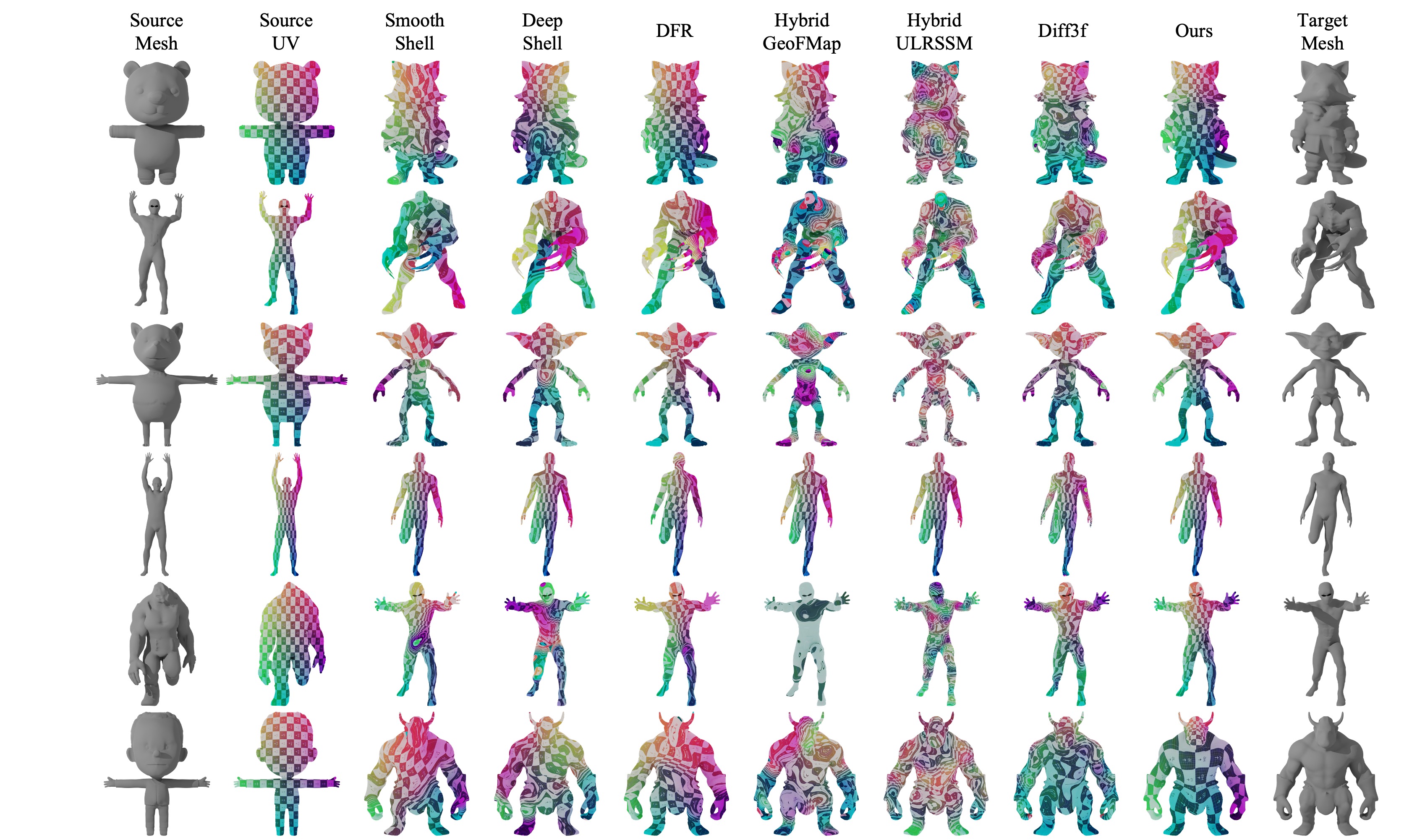}
  \caption{Visualized comparison with previous methods. All results are from the cross domain setting.}
  \label{fig_compare}
\end{figure*}

\subsection{Evaluation Metric and Dataset}

We adopt geodesic error normalized by the square root of the mesh’s total surface area as the evaluation metric.

To evaluate non-isometric shape correspondence, we introduce a new benchmark dataset, Character in-the-Wild (CharW), which includes 100 meshes from various sources, such as artists and 3D generative models~\cite{li2025triposg,zhang2024clay}. The CharW benchmark provides 3D correspondence annotations for evaluation, with further details in supplementary material. For evaluation, we use 3DBiCar~\cite{luo2023rabit} and SMPL~\cite{SMPL:2015,casas2023smplitex} as source meshes, and meshes from CharW as targets to form evaluation pairs. Additionally, we evaluate on the near-isometric FAUST remesh~\cite{bogo2014faust}, the non-isometric inter-class setting of DT4D-H~\cite{dt4d}, and CharW. From the FAUST test set, we sample 100 mesh pairs. DT4D-H includes two variants: DT4D-H std, the standard test set from~\cite{cao2023unsupervised}, and DT4D-H hard, a more challenging set that includes the Pumpkinhulk instance.

\subsection{Compared with previous methods\label{sec:compare}}
We compare our methods with previous methods such as registration method SmoothShell~\cite{smoothshell}, DeepShell~\cite{deepshell}, DFR~\cite{DeepFunctionalMapsPrior}, functional map series methods~\cite{donati2020deepGeoMaps,HybridFunctionalMaps,cao2023unsupervised} and Diff3F~\cite{Diff3f}. These methods are categorized by supervision type: (1) zero-shot methods that require no training, (2) unsupervised methods trained without ground-truth correspondence, and (3) supervised methods that rely on ground-truth annotations. Our full method is categorized as a supervised method. A discussion of this design choice is provided in supplementary material. %Sec.~\ref{sec:discussion}.\\
We evaluate all methods under two settings: (1) a cross-domain setting, where models are trained on 3DBiCar~\cite{luo2023rabit} and SURREAL~\cite{varol17_surreal}, and tested on other benchmarks to assess generalization; and (2) an intra-domain setting, where training and testing are on the same dataset, following standard practice in prior work. We introduce the cross-domain setting since it is more practical and able to generalize to various scenario. 
Quantitative results are shown in Table~\ref{tab_compare}, with visual comparisons in Figure~\ref{fig_compare}. Our method achieves state-of-the-art performance on non-isometric correspondence across all benchmarks. For isometric cases, we achieve the second-best result, on par with the current leading method~\cite{HybridFunctionalMaps}. Compared to Diff3F~\cite{Diff3f}, our results highlight that leveraging foundation model features alone is insufficient. By re-purposing foundation models for 2D character correspondence and introducing Semantic Flow Guided Registration, our method significantly improves performance. Additionally, our use of RGB images as input demonstrates strong robustness and flexibility across both textured and textureless scenarios.

\begin{table}[h!]
    \centering
    \caption{Ablation study on different design choices of our method. We use RGB images as inputs when evaluating on our Character in-the-wild benchmark dataset.}
    \resizebox{0.49\textwidth}{!}{
    \begin{tabular}{l|c|c|c}
        \toprule
        & \multicolumn{3}{|c}{Cross Domain} \\
        \textbf{Test on} & FAUST & CharW (RGB) & DT4D-H hard \\
        \midrule
        baseline& 2.88  & 3.44  & 6.04  \\
        Use Neural Jacobian Field  & 2.32  & 2.69  & 4.58  \\
        + shear-resistant loss  & 2.07  & 2.59  & 4.50  \\
        + geometry-grounded negative loss (full) & \textbf{1.83}  & \textbf{2.57}  & \textbf{4.12}  \\
        \bottomrule
    \end{tabular}}
    \label{tab_ablation_1}
\end{table}
\begin{table}[h!]
    \centering
    \vspace{-0.3cm}
    \caption{Ablation study on using feature adaptor.}
    \resizebox{0.45\textwidth}{!}{
    \begin{tabular}{l|c|c|c}
        \toprule
        & \multicolumn{3}{|c}{Cross Domain} \\
        \textbf{Test on} & FAUST & CharW (normal) & DT4D-H hard\\
        \midrule
        Diff3f~\cite{Diff3f}& 12.0  & 12.5  & 22.7  \\
        Ours (zero-shot)  & 5.60  & 3.48  & 14.1  \\
        Ours w/ feature adaptor & \textbf{1.83} & \textbf{2.61} & \textbf{4.12} \\
        \bottomrule
    \end{tabular}}
    \label{tab_ablation_adapter}
    \vspace{-0.3cm}
\end{table}
\begin{table}[h!]
    \centering
    \caption{Ablation study of different types of supervision given during registration.}
    \resizebox{0.45\textwidth}{!}{
    \begin{tabular}{l|c|c|c}
        \toprule
        & \multicolumn{3}{|c}{Cross Domain} \\
        \textbf{Test on} & FAUST & CharW (RGB) & DT4D-H hard\\
        \midrule
        3D correspondence supervision & 2.19  & 2.65  & 4.56  \\
        Semantic flow supervision& \textbf{1.83}  & \textbf{2.57}  & \textbf{4.12} \\
        \bottomrule
    \end{tabular}}
    \vspace{2mm}
    \label{tab_ablation_3dcorr}
\end{table}

\subsection{Ablation study}
We first conduct ablation studies to evaluate the effectiveness of three key components: (1) the Neural Jacobian Field, (2) the shear-resistant loss, and (3) the geometry-grounded negative loss used during 2D correspondence training. Our baseline is a simple deformation model that optimizes per-vertex displacements with Laplacian smoothing, without incorporating any of the above designs. As shown in Table~\ref{tab_ablation_1}, the Neural Jacobian Field significantly outperforms the per-vertex displacement baseline. Both the shear-resistant loss and the geometry-grounded negative loss contribute to improved correspondence quality.

Furthermore, we evaluate the necessity of training a 2D correspondence model with a feature adapter. For comparison, we construct a zero-shot variant that directly uses pre-trained Stable Diffusion and DINO features for 2D correspondence. As shown in Table~\ref{tab_ablation_adapter}, training a feature adapter significantly boosts performance, highlighting its importance.

We also examine the impact of using semantic flow as supervision for registration. To this end, we construct a baseline inspired by Diff3F~\cite{Diff3f}, where 2D features from the feature adapter network are first back-projected onto mesh vertices. Initial 3D correspondences are then obtained via nearest-neighbor search and used to supervise the registration. Results in Table~\ref{tab_ablation_3dcorr} demonstrate that using 2D semantic flow as guidance yields superior performance.

\begin{figure}[htb]
  \centering
  \includegraphics[width=.98 \linewidth]{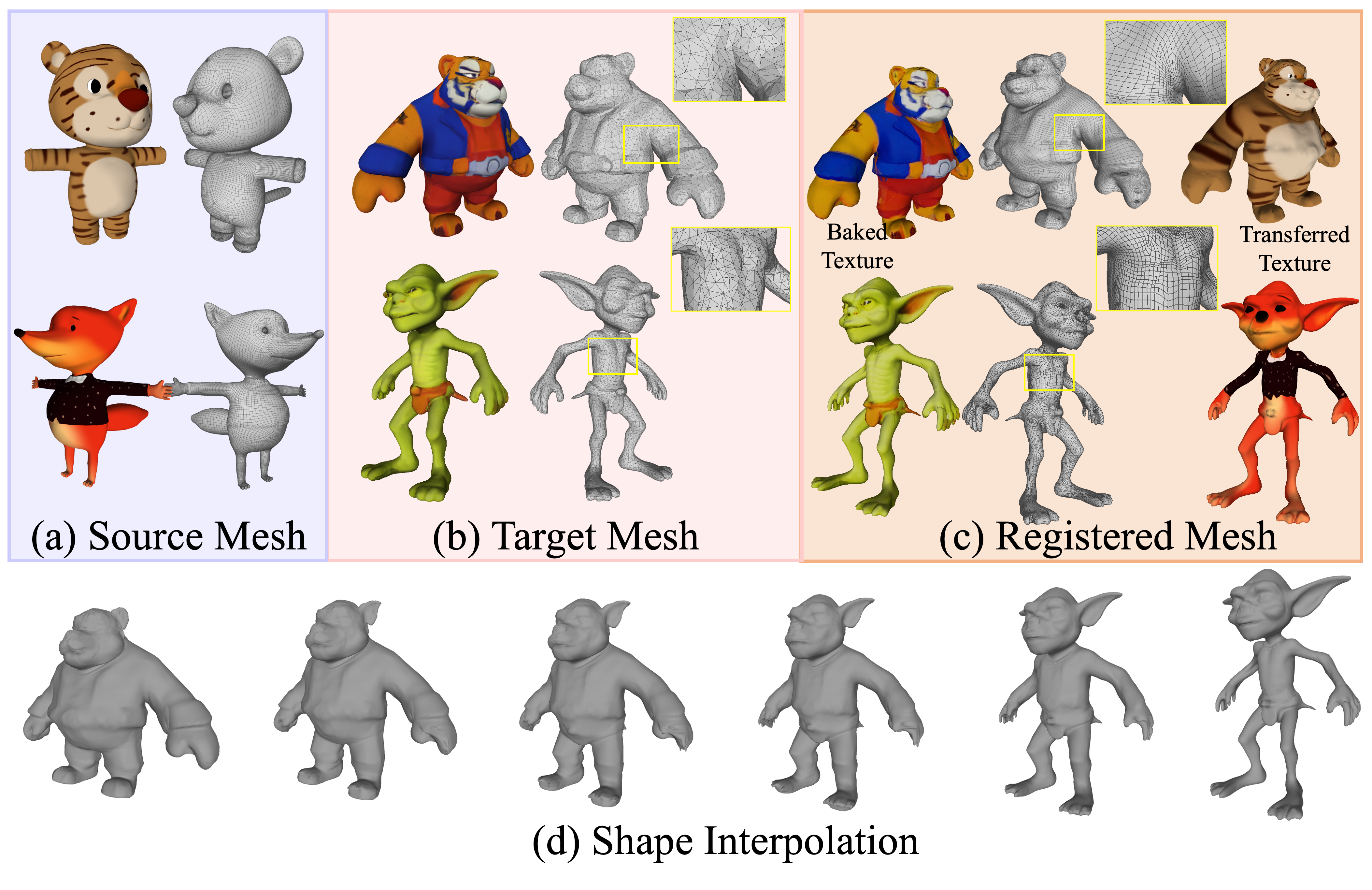}
  \caption{Application of re-topology, texture transfer and shape interpolation.}
  \label{fig_registration_morphing}
\end{figure}

\subsection{Downstream Application}
In this section, we demonstrate several downstream applications of Stable-SCore, including re-topology, texture transfer, and shape interpolation using characters generated by 3D generative models~\cite{li2025triposg,zhang2024clay,zhao2025hunyuan3d}. As shown in Figure~\ref{fig_registration_morphing}, we select source meshes from the 3DBiCar~\cite{luo2023rabit} dataset, which feature high-quality topology and artist-designed UV maps. We then apply Stable-SCore to register these source meshes to target meshes. The resulting registered meshes adopt the target’s shape while preserving the source’s topology and UV coordinates, enabling seamless texture transfer (Figure~\ref{fig_registration_morphing}(c)). Moreover, by registering 3DBiCar meshes to two arbitrary target meshes, we establish dense correspondences between them and perform shape interpolation (Figure~\ref{fig_registration_morphing}(d)).
\textit{Additional applications—including rig transfer and animation of characters generated by text-to-3D methods—are presented in the video in the \href{https://haolinliu97.github.io/Stable-Score/}{project page}.}
%\input{tables/tab_ablation_2}
% visualizaion:
% shape, pose, topology variations
% dataset diversity

% re-topology
% animation
% txture transfer
% general classes
\section{Conclusion}
We propose Stable-Score, a novel registration-based framework for stable 3D shape correspondence, addressing the challenge of non-isometric character matching. By adopting a 2D foundation model for robust 2D correspondences, our method guides a smooth registration process for accurate 3D alignment. Experiments show that Stable-Score significantly outperforms prior methods and enables downstream applications such as retopology, texture transfer, rig transfer, and shape interpolation. We also introduce the Character in-the-Wild (CharW) benchmark, a diverse dataset to further advance research in non-isometric correspondence. Our work pushes the state of 3D shape correspondence forward, opening new opportunities for both academic and practical applications.

\section{Acknowledgements}
The work was supported in part by the Basic Research Project No.~HZQB-KCZYZ-2021067 of Hetao Shenzhen-HK S\&T Cooperation Zone, by Guangdong Provincial Outstanding Youth Fund (No.~2023B1515020055), by Shenzhen Science and Technology Program No.~JCYJ20220530143604010, and No.~NSFC\-61931024. 
It is also partly supported by the National Key R\&D Program of China with grant No.~2018YFB1800800, by Shenzhen Outstanding Talents Training Fund 202002, by Guangdong Research Projects No.~2017ZT07X152 and No.~2019CX01X104, by Key Area R\&D Program of Guangdong Province (Grant No.~2018B030338001), by the Guangdong Provincial Key Laboratory of Future Networks of Intelligence (Grant No.~2022B1212010001), and by Shenzhen Key Laboratory of Big Data and Artificial Intelligence (Grant No.~ZDSYS201707251409055).  
{
    \small
    \bibliographystyle{ieeenat_fullname}
    \bibliography{main}
}
\clearpage
\section{Character in-the-wild Benchmark}
\label{sec:charw}
\begin{figure}[htb]
  \centering
  \includegraphics[width=.98 \linewidth]{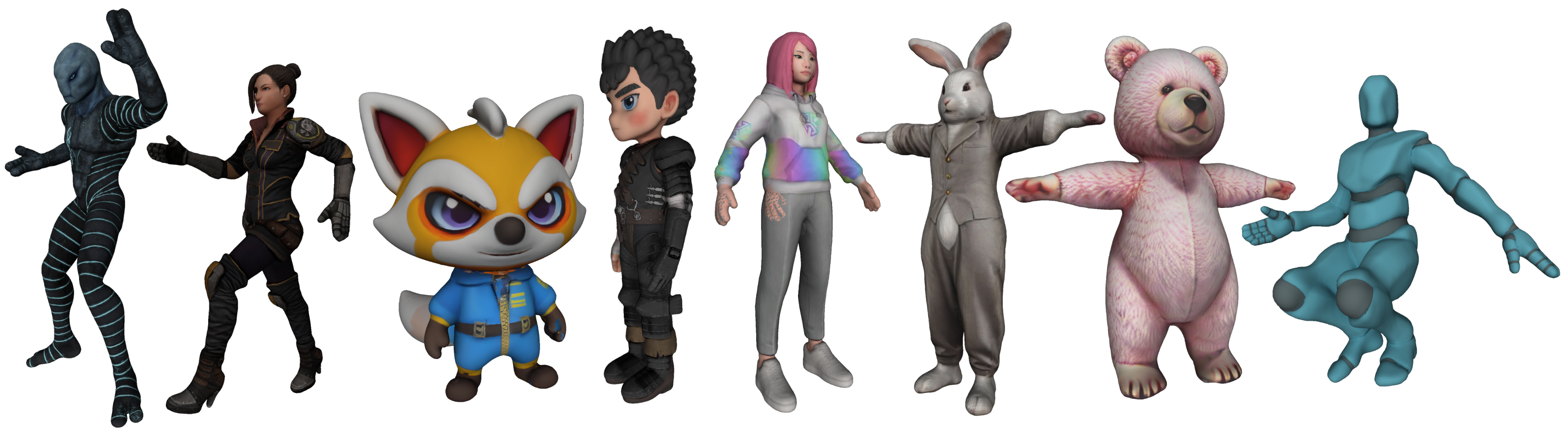}
  \caption{Visualized examples of the diverse shapes in our \textit{Character in-the-wild} benchmark.}
  \label{fig_benchmark}
\end{figure}

To better evaluate non-isometric shape correspondence, we introduce a new benchmark dataset, Character in-the-Wild (CharW). It consists of 100 character meshes collected from various sources, including artists and 3D generative models~\cite{li2025triposg,zhang2024clay}. Dense correspondences are annotated by manually deforming template meshes from 3DBiCar~\cite{luo2023rabit} and SMPL~\cite{SMPL:2015} to align with the collected shapes. By using a shared set of templates, ground-truth correspondences can be established between any pair of meshes in the benchmark. We compare CharW with several public character correspondence benchmarks~\cite{bogo2014faust,anguelov2005scape,melzi2019shrec,lahner2016shrec,dt4d} in Table~\ref{table:benchmark}. Unlike previous datasets, which are typically re-meshed versions of original mesh, CharW features greater diversity in shape and topology. Visualization examples are shown in Figure~\ref{fig_benchmark}. 

% We further produce dense correspondence annotations by deforming 3DBiCar~\cite{luo2023rabit} or SMPL~\cite{SMPL:2015} meshes to manually align them with the collected meshes, done by professional artists. 
%Since 3DBiCar and SMPL are parametric models and all meshes share the same vertices order, correspondence between any two meshes from \textit{CharW} can be retrieved. 

\begin{table}[h]
\caption{Quantitative comparison between our \textit{Character in-the-wild} benchmark dataset and other character correspondence benchmark dataset.}
\resizebox{0.99\linewidth}{!}{
\begin{tabular}{c|c|c|c|c}
\toprule
\textbf{Dataset} & \# of identities & shape variance &Non-isometric & various sources\\
\midrule
FAUST~\cite{bogo2014faust} & 10 & small & &  \\
SCAPE~\cite{anguelov2005scape} & 1 & small & & \\
SHREC'19~\cite{melzi2019shrec} & 44 & small & & \\
TOPKIDS~\cite{lahner2016shrec} & 1 & small& & \\
DT4D-H~\cite{dt4d} & 10 & large & \checkmark & \\
CharW(Ours) & 100 & large & \checkmark & \checkmark \\
\bottomrule
\end{tabular}
}
\label{table:benchmark}
\end{table}

\subsection{CharW Dataset Curation Details}
\paragraph{Data Collection}

We collect meshes from two main sources: generated meshes and artist-crafted meshes. The generated meshes come from Tripo~\cite{li2025triposg} and Rodin~\cite{zhang2024clay}, while the artist-designed meshes are sourced from Mixamo and Sketchfab. The dataset includes 38 meshes from Mixamo, 23 from Rodin, 22 from Tripo3D, and 40 from Sketchfab. Several criteria were followed during data collection:
	1.	All meshes represent bipedal characters, including both human and humanoid figures.
	2.	The meshes must be of high quality, free of significant distortion or artifacts, and avoid low-poly models.
	3.	We selected characters with diverse body types such as fat, slim, large-headed, and small-headed, to ensure a variety of non-isometric shapes.

\paragraph{Correspondence Annotation}

CharW benchmark provides correspondence annotations for evaluation. The annotation process begins by selecting the SMPL neutral template mesh and three meshes from 3DBiCar~\cite{luo2023rabit} as deformation templates. For each target mesh, a professional artist first selects the template that most closely resembles the target. The artist then annotates key-point correspondences between the source and target meshes, ensuring that the key points are semantically aligned. On average, 60 to 80 key points are annotated per mesh. After annotation, the artist uses ZBrush’s warp add-on to align the source mesh with the target. The annotated key points are refined until the results are satisfactory. If needed, the artist manually adjusts the warped mesh using sculpting tools. Finally, the warped mesh is processed with Blender’s Shrinkwrap modifier to precisely match the target.

\section{Supplementary}
\subsection{Limitation and Future work}
There are several limitations stemming from the 2D correspondence model: (1) significant initial rotation misalignment, which leads to incorrect 2D correspondences and distorts the deformation process; (2) severe occlusion, resulting in missing 2D correspondences; and (3) difficulty handling complex structures due to the low resolution of 2D features, making it challenging to capture details such as fingers, as shown in the first row of Fig~\ref{fig:charw1}.

The deformation process also has limitations: (1) topological noise, such as the body and arms merging together, hinders deformation; (2) it inherits limitations from NJF, including the inability to handle partial shapes.
An interesting direction for future work is to move beyond the limitations of low-resolution 2D prior models by integrating native 3D prior models, such as large 3D generative models~\cite{zhang2024clay, li2025triposg, zhao2025hunyuan3d}, which could potentially address these issues. \\

\subsection{Discussion on Supervision Types}
\label{sec:discussion}
Previous functional map-based methods~\cite{DeepFunctionalMapsPrior,DeepGeometricFunctionalMaps,sun2023spatially,li2024deformable,attaiki2023shape,attaiki2023understanding,magnet2023scalable,ovsjanikov2012functional,donati2020deepGeoMaps} are mostly unsupervised. Supervised methods~\cite{DeepGeometricFunctionalMaps} are rare in shape correspondence, and previous supervised attempts have underperformed compared to unsupervised methods. While supervised methods generally outperform unsupervised ones in most computer vision tasks, this is not the case for shape correspondence. We hypothesize that design flaws in earlier supervised methods, such as strong reliance on LBO’s basis or DiffusionNet’s prior~\cite{sharp2022diffusionnet}, lead to unsatisfactory results and overfitting problems, suggesting significant room for improvement.

To address this, we propose a supervised, registration-based method that avoids reliance on functional maps or DiffusionNet. Our approach achieves state-of-the-art performance on various non-isometric benchmarks. Although it requires ground-truth (GT) supervision, the annotation cost is relatively low, requiring only sparse 2D/3D key points or dense correspondences via template warping (same as the CharW annotation process). This makes our method highly cost-effective, therefore, the need for GT supervision is no longer a limitation compared to unsupervised approaches.

% \subsection{Evaluation on Standard DT4D testset}
% We also includes comparison on Standard DT4D test set as used in \cite{cao2023unsupervised,HybridFunctionalMaps}, as shown in table ~\ref{tab_compare:dt4d-std}. Our method still outperform other methods under the standard DT4D test set.

% \begin{table}[h!]
%     \centering
%     \caption{Quantitative comparison between our methods and previous methods on standard DT4D test set}
%     \resizebox{0.40\textwidth}{!}{
%     \begin{tabular}{l|c|c}
%         \toprule
%         & {Cross Domain} & {Intra Domain} \\
%         \midrule
%         SmoothShell~\cite{smoothshell} & 13.59 & 13.59  \\
%         Diff3f~\cite{Diff3f} & 24.01 & 24.01\\
%         \midrule
%         DeepShell~\cite{deepshell} & 30.96 & 29.14\\
%         DFR\cite{DeepFunctionalMapsPrior} & 19.83 & 14.94\\
%         HybridGeoFMap~\cite{HybridFunctionalMaps} & 22.12 & 4.08  \\
%         ULRSSM~\cite{cao2023unsupervised} & 28.2 & 4.61\\
%         Hybrid ULRSSM~\cite{HybridFunctionalMaps} & 15.46 & 3.47\\
%         \midrule
%         GeoFMap~\cite{donati2020deepGeoMaps}  & 25.15 & 4.12\\
%         Ours (Normal) & \textbf{4.23} & \textbf{3.11}\\
%         \bottomrule
%     \end{tabular}}
%     \label{tab_compare:dt4d-std}
% \end{table}

\subsection{Beyond Characters} 
\begin{figure}[h]
    \centering
    \includegraphics[width=0.99\linewidth]{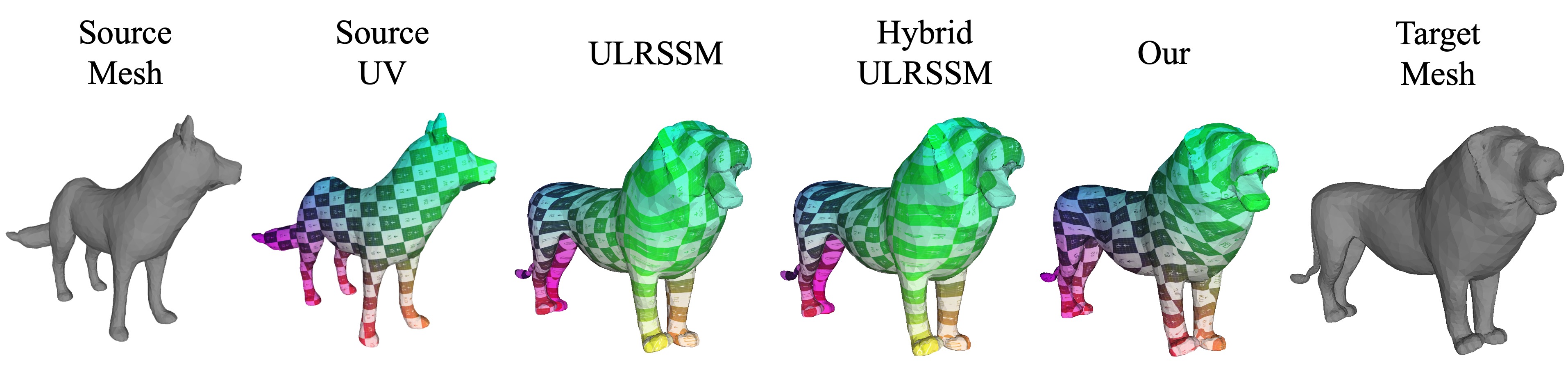}
    \caption{\small{Comparison on SMAL, all methods are trained on SMAL.}}
    \label{fig:SMAL}
\end{figure}

We also test the applicability of our method to other domains, such as animal shape correspondence, which is commonly evaluated in non-isometric shape matching. Additionally, we compare our method on the SMAL dataset, as shown in Table~\ref{tab_compare:smal} and Figure~\ref{fig:SMAL}. The quantitative comparisons are conducted under two setups: training on a character dataset or on the SMAL dataset, with testing performed on the SMAL dataset. “Ours (zero-shot)” refers to a zero-shot version of our method, where feature adapters are removed, and no further fine-tuning is required. Our method outperforms others and demonstrates its ability to generalize to domains where ground truth correspondences are available for training. Furthermore, the zero-shot setup of our method is versatile, achieving strong performance across various tasks.

\begin{table}[h!]
\centering
\caption{\label{tab_compare:smal}Quantitative comparison between our methods and previous methods on SMAL dataset.}
\resizebox{0.40\textwidth}{!}{
\begin{tabular}{c|c|c}
\toprule
Test on SMAL& Train on character & Train on SMAL\\
\midrule
ULRSSM & 28.5 & 3.63\\
Hybrid ULRSSM & 44.0 & 3.11\\
Ours (zero-shot) & \textbf{8.91} & 8.91\\ 
Ours (full) & 17.01 & \textbf{2.65} \\
\bottomrule
\end{tabular}
}
\end{table}

\section{More results}
\begin{figure*}[ht]
\includegraphics[width=.99\linewidth]{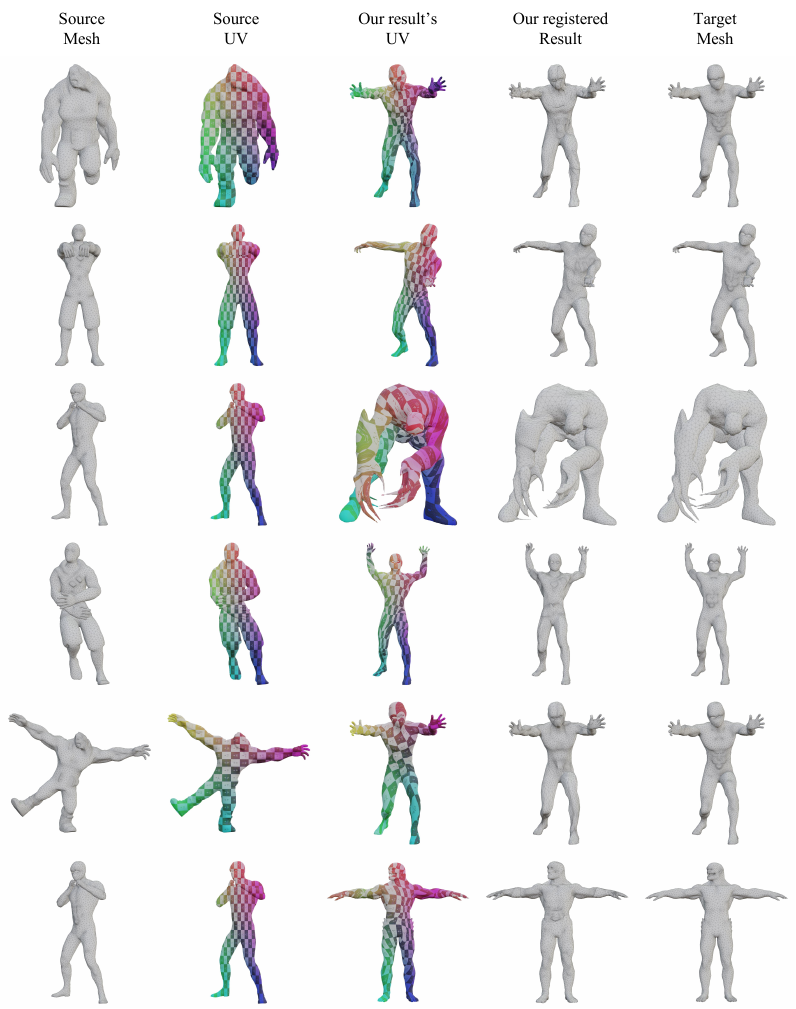}
\caption{Visualized results produced by Stable-Score on DT4D dataset.\label{fig:DT4D}}
\end{figure*}

\begin{figure*}[ht]
\includegraphics[width=.99\linewidth]{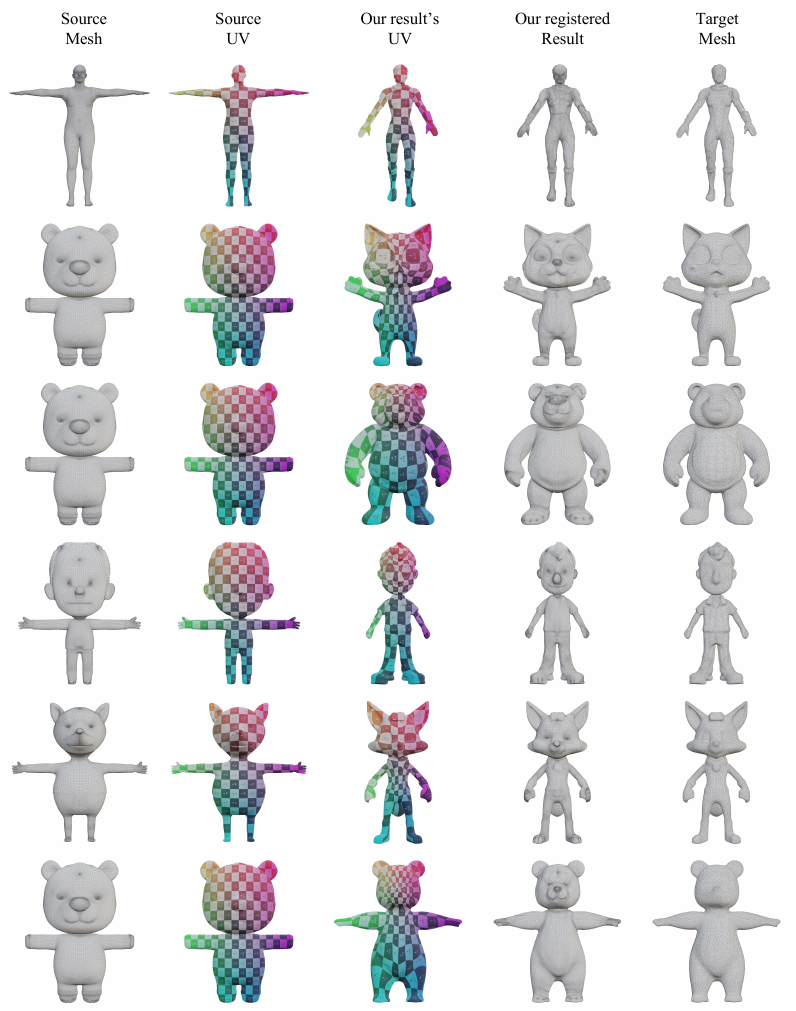}
\caption{Visualized results produced by Stable-Score on our CharW dataset.\label{fig:charw1}}
\end{figure*}

\begin{figure*}[ht]
\includegraphics[width=.99\linewidth]{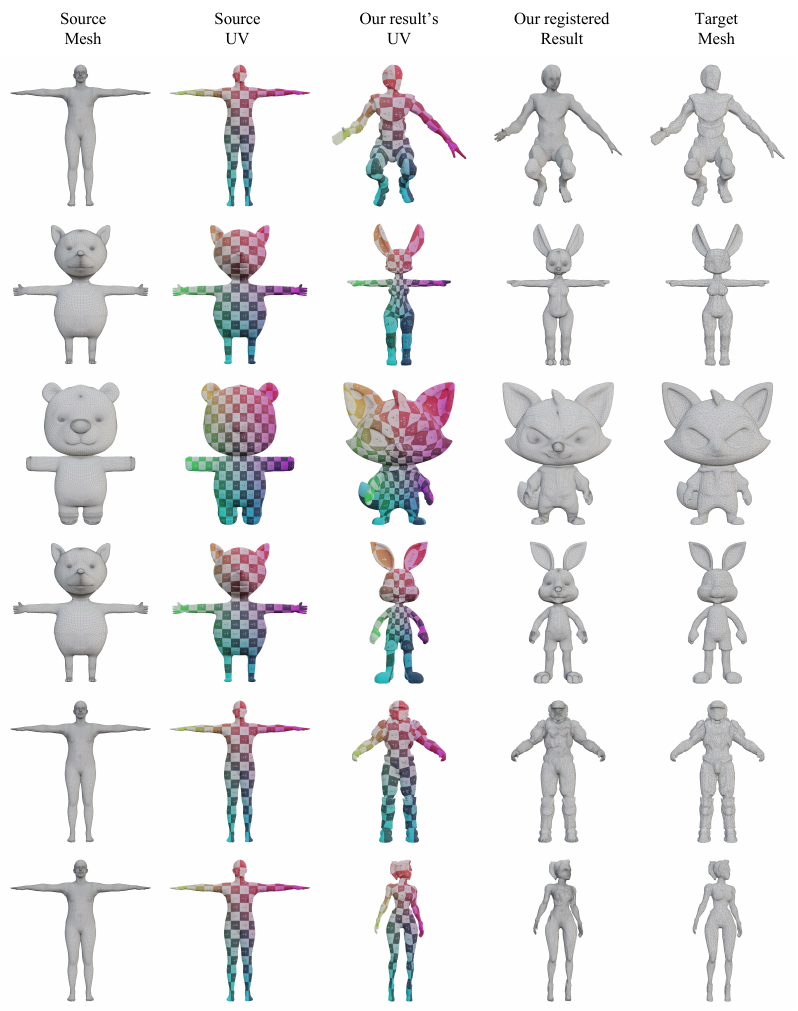}
\caption{Visualized results produced by Stable-Score on our CharW dataset.\label{fig:charw2}}
\end{figure*}

\begin{figure*}[ht]
\includegraphics[width=.99\linewidth]{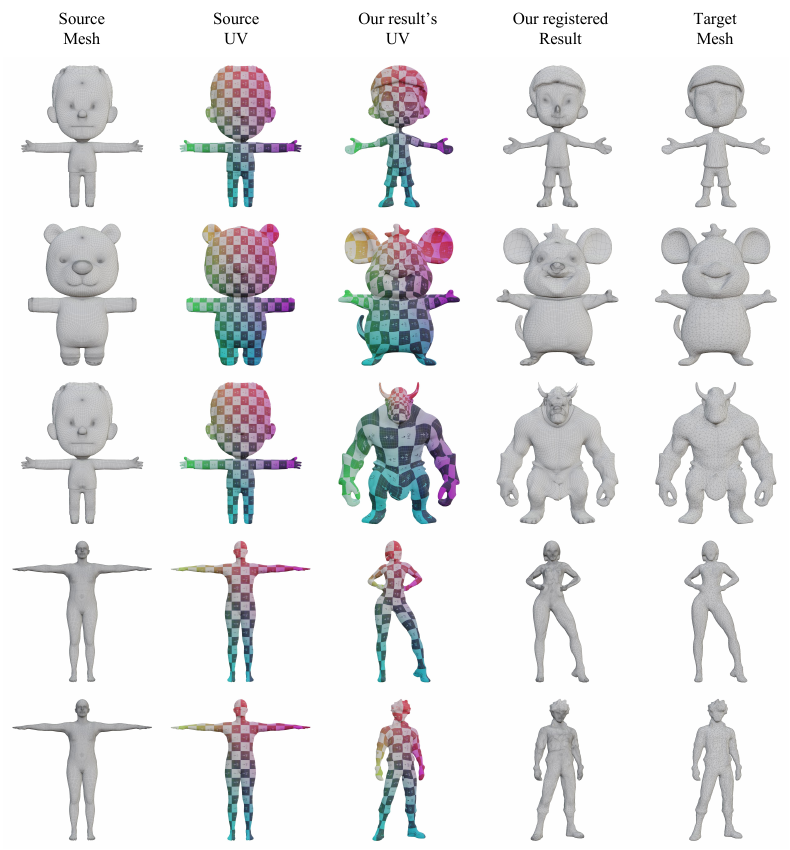}
\caption{Visualized results produced by Stable-Score on our CharW dataset.\label{fig:charw3}}
\end{figure*}

More results on the DT4D dataset and CharW dataset are shown in Figure~\ref{fig:DT4D}, \ref{fig:charw1}, \ref{fig:charw2} and \ref{fig:charw3}. It can be observed that our Stable-Score achieves precise registration while simultaneously preserving the source topology, thereby demonstrating significant potential for re-topology applications.



\end{document}